\documentclass[11pt]{article}

\usepackage[final]{acl}

\usepackage{times}
\usepackage{latexsym}

\usepackage[T1]{fontenc}

\usepackage[utf8]{inputenc}

\usepackage{microtype}

\usepackage{inconsolata}

\usepackage{graphicx}
\usepackage{enumitem}
\usepackage{booktabs} 
\usepackage{multirow}
\usepackage{adjustbox}
\usepackage{colortbl}
\usepackage{subfig}
\usepackage{amsmath}
\usepackage{xcolor}
\usepackage{amssymb}
\usepackage{mathtools}
\usepackage{tcolorbox}
\usepackage{caption}     
\usepackage{siunitx} 

\definecolor{mixedred}{RGB}{255, 100, 100}
\newcommand{\gradcell}[2][]{%
  \cellcolor{mixedred!#1}
  #2
}

\title{Why Did Apple Fall: Evaluating Curiosity in Large Language Models}

\author{
  \textbf{Haoyu Wang\textsuperscript{1}} 
  \textbf{Sihang Jiang\textsuperscript{1}} 
  \textbf{Yuyan Chen\textsuperscript{3}}
  \textbf{Xiaojun Meng\textsuperscript{2}}
  \textbf{Jiansheng Wei\textsuperscript{2}}
\\
  \textbf{Yitong Wang\textsuperscript{1,*}}
  \textbf{Yanghua Xiao\textsuperscript{1,*}}
\\
\\
  \textsuperscript{1}Shanghai Key Laboratory of Data Science, School of Computer Science, Fudan University
\\
  \textsuperscript{2}Huawei Large Model Data Technology Lab,
  \textsuperscript{3}Cornell University
\\
  \small\texttt{\{wanghy24@m., yitongw@, shawyh@\}fudan.edu.cn}
}

\begin{document}
\maketitle
\begin{abstract}
Curiosity serves as a fundamental construct in human cognition.      
Inspired by curiosity, reinforcement learning with intrinsic rewards for large language models (LLMs) has shown substantial potential.      
However, it remains unclear whether existing curiosity-driven methods genuinely reflect curiosity-like behaviors in LLMs, and to what extent psychological notions of curiosity can be transferred to these models.
In this work, we propose a psychology-inspired framework to evaluate and leverage curiosity in LLMs.      
We adapt the Five-Dimensional Curiosity scale Revised (5DCR) to LLMs and combine questionnaire-based self reports with behavioral study.      
We find that although LLMs can exhibit curiosity-like behavioral patterns resembling those of humans, such patterns do not reflect an intrinsic trait of curiosity.    
Building on this insight, we design a curiosity-driven thinking pipeline to examine the functional role of human-like curious behaviors.    Experiments show that instructing LLMs to emulate curious strategies leads to better performance on selected downstream tasks, indicating that mimicking curious behaviors holds promise for reasoning enhancement.\footnote[1]{Code: https://github.com/Yukijudaii1352/CuriosityEval}
\end{abstract}

\section{Introduction}

\begin{figure}[!ht]
  \centering
  \includegraphics[width=0.9\linewidth]{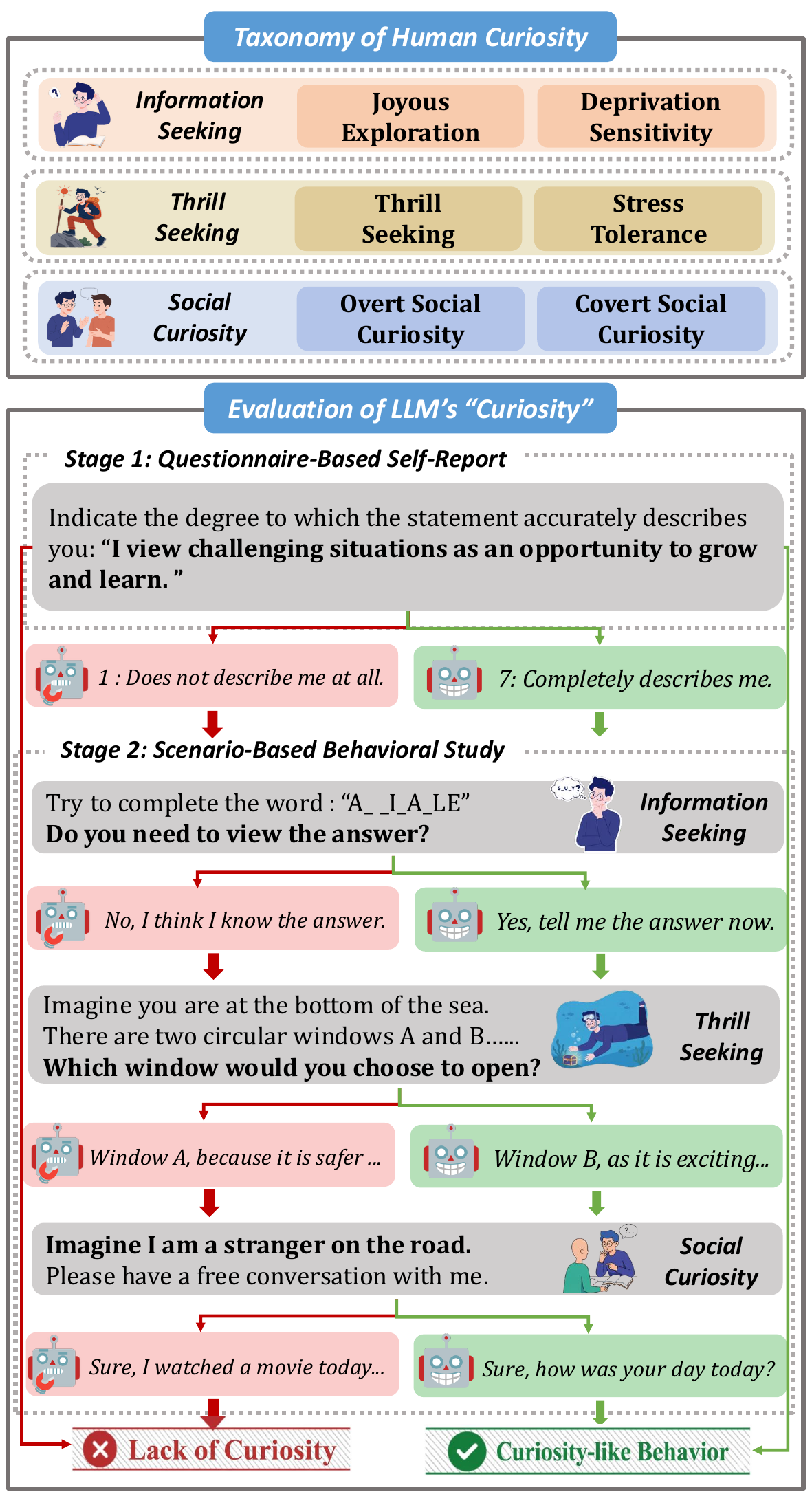}
  \caption{Taxonomy of human curiosity and our two-stage framework for evaluating curiosity-like behavior of LLMs, combining questionnaire-based self-report with scenario-based behavioral tasks across information seeking, thrill seeking, and social curiosity.}
  \label{fig:overview}
  \vspace{-1em}
\end{figure}

\begin{flushright}
  \begin{minipage}{1.2\linewidth}  
    \begin{quote}
      ``\textit{I have no special talent. \\
      I'm only passionately curious.}'' 
    \end{quote}
  \end{minipage}
\end{flushright}
\begin{flushright}
  \begin{minipage}{0.6\linewidth} 
    \begin{quote}
        \hfill — \underline{Albert Einstein}  
    \end{quote}
  \end{minipage}
\end{flushright}


\textbf{Curiosity}, an intrinsic human trait that emerges when facing novel environments, plays a crucial role in the discovery and acquisition of knowledge \cite{berlyne1954theory,loewenstein1994psychology}. Numerous scientific breakthroughs have been spurred by curiosity, as exemplified by Sir Isaac Newton's formulation of universal gravitation following the observation of a falling apple. Given the central role that curiosity plays in human learning, it is natural to ask whether analogous mechanisms can be instantiated in artificial systems such as large language models (LLMs).

A recently popular training paradigm for LLMs is reinforcement learning with verifiable rewards (RLVR), which directly optimizes models using verifiable signals of correctness derived from final answers. Representative methods within this paradigm include GRPO \cite{shao2024deepseekmath} and DAPO \cite{yu2025dapo}. Within this framework, several studies have proposed intrinsic reward signals inspired by human curiosity, such as i-MENTOR \cite{gao2025navigate} and CDE \cite{dai2025cde}, aiming to guide models toward more autonomous exploration and learning behaviors. These approaches generally encourage action trajectories that are more uncertain or less frequently explored, thereby providing a behavioral analogy to human curiosity, and have shown potential in math and coding. 

However, a fundamental question whose answer measures the validity of existing research in this line has not received enough attention: \textbf{\textit{Do LLMs possess curiosity?}} Current curiosity-driven RL methods face two key issues. First, it has not been adequately evaluated whether LLMs can exhibit behavioral characteristics that are meaningfully analogous to curiosity.  Second, many such methods rely on amplifying statistical signals such as entropy \cite{agarwal2025unreasonable} and perplexity \cite{cui2025stepwise}. This makes it difficult to disentangle whether improvements stem from strengthened supervision signals or from genuinely curiosity-like behavior exhibited by the model.  Therefore, it is necessary to systematically evaluate curiosity-related behaviors in LLMs and to analyze the benefits attributable to such behaviors themselves, thereby providing an empirical basis for understanding the scope of the concept of curiosity in LLMs and its relationship to intrinsic rewards.

To address these questions, we propose a psychology-inspired framework to systematically evaluate and leverage curiosity in LLMs, as illustrated in Figure~\ref{fig:overview}.
(A) \textbf{\textit{Taxonomy of  curiosity}}. We first build a conceptual foundation by adapting the Five-Dimensional Curiosity scale Revised (5DCR) \cite{kashdan2020five} to the LLM setting, consolidating it into three core dimensions:Information Seeking, Thrill Seeking, and Social Curiosity.
(B) \textbf{\textit{Questionnaire-based self-report}}. Based on this taxonomy, we then elicit self-reported curiosity profiles by prompting LLMs with 5DCR-derived questionnaires, yielding an initial, questionnaire-level and introspective assessment of their curiosity.
(C) \textbf{\textit{Scenario-Based Behavioral Study}}. We next transcend self-report by designing corresponding behavioral tasks for each dimension. For each dimension in part (A), we adapt corresponding behavioral decision-making tasks and evaluate the models’ choices in these scenarios, thereby systematically testing whether the curiosity traits inferred from questionnaires in part (B) are consistently reflected in actual behavior.
(D) Finally, leveraging insights from these analyses, we construct a curiosity-driven questioning-and-thinking pipeline that explicitly instructs LLMs to emulate curiosity-like exploratory behaviors, enabling us to probe, to quantify, and to enhance how curiosity-inspired behavior affects their downstream reasoning and problem-solving performance.

To the best of our knowledge, this is the first work that systematically evaluates curiosity-related behaviors in LLMs. Our findings indicate that:

\noindent \textit{1. LLMs display consistent curiosity-like patterns across questionnaire-based and behavioral evaluations.} 
The alignment between self-report and scenario-based choices suggests that psychological instruments can be adapted to systematically assess curiosity-like tendencies in LLMs.

\noindent \textit{2. The “curiosity” of LLMs is asymmetric and highly context-sensitive: strong information seeking but limited thrill seeking.} 
Meanwhile, these curiosity-like patterns are often unstable across prompts, accompanied by low verbal confidence, and highly sensitive to contextual variations. 
Taken together, they are more naturally interpreted as artifacts of fitting human behavioral data and safety constraints rather than evidence of a genuine intrinsic exploratory drive.

\noindent \textit{3. Human-like curious questioning  can  be harnessed to improve LLM performance.} 
When we explicitly prompt models to emulate human-style curiosity by encouraging auxiliary questioning and reflective thinking, they tend to generate higher-quality intermediate thinking, and achieve better performance on selected downstream tasks. 
This indicates that even purely behavioral analogs of curiosity, instead of modifying intrinsic objectives, can be practically leveraged to enhance reasoning and problem-solving in LLMs.




\section{Related Work}

\subsection{Evaluation of Curiosity}
Research and evaluation of human curiosity originated from Berlyne's four-dimensional model of curiosity \cite{berlyne1966curiosity}, which categorizes curiosity into perceptual curiosity, cognitive curiosity, trait curiosity, and state curiosity.  As definitions and descriptions of curiosity became more refined, the assessment of curiosity shifted towards scale-based methods, with significant contributions from the Perceptual Curiosity Scale (PC) \cite{collins2004measurement} and the Curiosity and Exploration Inventory (CEI) \cite{kashdan2009curiosity}. The most recent and authoritative scale in psychological research for measuring curiosity is the Five-Dimensional Curiosity Scale Revised (5DCR) \cite{kashdan2020five}, which based on a six-dimensional model of curiosity and refines the wording of certain items. Consequently, this paper adopts the 5DCR and its dimensional definitions of curiosity as the theoretical reference.

\subsection{Curiosity in Language Model}
The theory of curiosity has been widely applied in natural language processing research, with its initial adoption in question generation and evaluation.  Scialom et al. assessed the degree of curiosity in questions within the SQuAD dataset based on BLEU scores and perplexity \cite{scialom2020ask}.  Gao et al. designed multiple types of curious questions, which were annotated by humans, and demonstrated that curious questions contribute to improvements in the grammaticality, semantics, and relevance of generated texts \cite{gao2022makes} .  Javaji et al. collected questions posed by LLMs in disciplines such as physics and chemistry, evaluating the models’ abilities to generate curious questions \cite{javajiwould}.  In recent years, curiosity-driven RL algorithms have also achieved notable success.  CD-RLHF enhances the diversity of texts generated by LLMs by rewarding the generation of less common tokens \cite{sun2025curiosity}. CDE formalizes curiosity through signals from both the actor and critic, which helps to mitigate premature convergence and collapse in the reasoning processes of models \cite{dai2025cde}.

\section{Method}
In this section, we provide a detailed explanation of how to evaluate the curiosity-like behavior of LLMs through questionnaires and behavioral experiments, and how to enhance the learning ability of LLMs through curiosity-driven questioning. The framework is shown in Figure \ref{fig:overview}. Detailed experimental settings are provided in the Appendix.

\subsection{Curiosity Questionnaire Study}
Initially, adhering to established methodologies for assessing human curiosity, we prompt LLMs to self-evaluate their level of curiosity using the Five-Dimensional Curiosity Scale Revised (5DCR) \cite{kashdan2020five}. The 5DCR conceptualizes human curiosity across three overarching dimensions—\textit{Information Seeking}, \textit{Thrill Seeking}, and \textit{Social Curiosity}—which are further delineated into six specific subdimensions: Joyous Exploration (\textbf{JE}), Deprivation Sensitivity (\textbf{DS}), Stress Tolerance (\textbf{ST}), Thrill Seeking (\textbf{TS}), Overt Social Curiosity (\textbf{OSC}), and Covert Social Curiosity (\textbf{CSC}). The scale consists of 24 items, with each subdimension represented by four statements. For each statement, the model is prompted to rate the extent to which the statement reflects its own disposition using a seven-point Likert scale (1 = strongly disagree, 7 = strongly agree). 

In addition to mean and variance, we used several other metrics to compare model and human performance and assess the validity of the model’s self-reports. Cohen's d \cite{diener2010cohen} quantifies the standardized mean difference between models and humans:
$$
d = \frac{\bar{X}_1 - \bar{X}_2}{s}
$$
where $\bar{X}_1$ and $\bar{X}_2$ are sample means and $s$ is the pooled standard deviation.  

McDonald's Omega Coefficient \cite{bonniga2020literature} reflects the proportion of total variance attributable to common variance among observed variables, serving as an index of questionnaire reliability. Based on Confirmatory Factor Analysis (CFA) \cite{harrington2009confirmatory}, it is defined as:
$$
\omega = \frac{( \sum_{i=1}^{k} \lambda_i )^2}{( \sum_{i=1}^{k} \lambda_i )^2 + \sum_{i=1}^{k} \theta_i }
$$
where $k$ is the number of items, $\lambda_i$ is the standardized factor loading of the $i$‑th item on the common factor, and $\theta_i$ is its error variance. Higher $d$ indicates greater divergence between model and human performance, while higher $\omega$ indicates greater internal consistency of questionnaire responses.

\subsection{Curious Behavior Study}
The questionnaire-based responses of LLMs are not entirely reliable, as they can be influenced by factors such as hallucinated personas, item phrasing, and response scales \cite{han2025personality}. 
Therefore, for each of the three dimensions of curiosity, we adapt behavioral experiments from psychology that were originally designed for humans. 
These experiments allow us to validate the questionnaire-based findings by examining the models' actual choices in curiosity-related scenarios.

\textbf{Curiosity-Driven Information Seeking} 
We adapt the missing-letter game designed by Singh et al. \cite{singh2021effect} to evaluate Curiosity-Driven Information Seeking. 
The model is prompted to complete a series of incomplete words. 
After providing its guess for each word, the model can choose whether to reveal the correct answer. 
Because revealing the answer does not affect any external reward, the decision to request the answer primarily reflects an intrinsic desire to reduce uncertainty about its own prediction. 
Thus, the frequency with which the model chooses to see the correct answers serves as a behavioral proxy for information-seeking curiosity in the presence of knowledge gaps.

\textbf{Curiosity-Driven Thrill Seeking} 
We adapt the underwater exploration game proposed by Jirout et al. \cite{jirout2012children} to evaluate Curiosity-Driven Thrill Seeking. 
The LLM is placed in a role-play scenario where it is inside a submarine and can observe fish through porthole windows. 
On each trial, the model chooses between two windows that differ in outcome uncertainty: 
a low-uncertainty window, where the exact type of fish to be seen is specified in advance, and a high-uncertainty window, where the fish type is random and unpredictable. 
Since the options are matched in external payoff and differ mainly in uncertainty and novelty, a systematic preference for the high-uncertainty window can be interpreted as a form of thrill-seeking curiosity—a willingness to engage with more unpredictable and potentially more exciting outcomes. 

\textbf{Social Curiosity} 
In human research, the frequency and depth of questions asked in conversation are well-established proxies for social curiosity, as they reflect an active effort to learn about another person.
We adapt the conversational paradigm introduced by Hartung et al. \cite{hartung2011social} to evaluate Social Curiosity. 
The model engages in ten rounds of informal conversation with a virtual stranger who has a randomly assigned personality profile (role-played by Grok-3). 
Within these interactions, the model is free to respond, self-disclose, or ask questions. We use the frequency of questions posed by the model during these conversations as a proxy for its curiosity about other people in social contexts.

\subsection{Curiosity-Driven Learning}
\begin{table}[ht]
\centering
\small
\begin{tabular}{p{1.8cm}p{4.9cm}}
\toprule
\multicolumn{2}{c}{\textbf{Instruction}} \\  
\midrule
\textbf{Vanilla CoT} & 
Read the given context, then answer the question. When answering, think step by step first and then form the answer.\\
\midrule
\textbf{Refined CoT} & 
Read the given context, then answer the question. When answering, think step by step first and then form the answer. \textcolor{blue}{\textit{You are encouraged to add reflection and self-improvement, such as 'Wait, something is wrong here.' or 'I need to rethink this.'}}\\
\midrule
\textbf{Curious CoQ} &
\textcolor{red}{\textit{You are a smart and curious student}}. Read the given context, then answer the question. When answering, think step by step first and then form the answer. \textcolor{red}{\textit{You are encouraged to add some self-asking and self-answering questions when thinking, such as 'What if', 'Why', 'How' etc.}} \\
\bottomrule
\end{tabular}
\caption{System prompt for vanilla  and curiosity-driven learning of LLMs.}
\label{tab:prompt}
\end{table}

After evaluating curiosity-like behaviors in LLMs, we next explore a pragmatic question: Can we harness the benefits of curiosity without complex reward engineering? Previous works have sought to instill curiosity through specialized intrinsic rewards within RL frameworks. In contrast, we investigate a lighter-weight, more directly applicable alternative: eliciting curiosity-like behaviors purely through prompting or supervised fine-tuning. This approach tests whether the behavioral form of curiosity itself is beneficial, even in the absence of a dedicated curiosity-driven objective.

In humans, curiosity-driven questioning is a crucial component of learning, as it fosters exploration and knowledge acquisition \cite{balim2009effects}. 
Motivated by this analogy, we study whether explicitly encouraging curiosity-like questioning can improve the quality of intermediate chains-of-thought and downstream task performance in LLMs. The most direct method is persona injection \cite{chen2024persona}, which involves guiding the model to become more curious through prompt engineering.  We compare the effects of two types of prompts on model reasoning: Chain-of-Thought (CoT) \cite{wei2022chain} and Curious Chain-of-Questioning (CoQ), as detailed in Tab. \ref{tab:prompt}. Another approach to enable LLMs to exhibit curiosity similar to that of humans is the injection of supervision signals.  We compare the differences in reasoning abilities between models trained with CoT and CoQ paradigms.  Three distinct types of thinking processes are collected for training:

\noindent \textbf{Vanilla CoT}: The optimal reasoning process that directly leads to the answer.

\noindent \textbf{Refined CoT}: Reasoning processes that incorporate reflection and backtracking, characterized by "Aha Moments." \cite{guo2025deepseek}  These thinking processes are distilled from DeepSeek-R1, which has been demonstrated to improve model accuracy through fine-tuning.

\noindent \textbf{Curious CoQ}: Curiosity-driven, question-centric reasoning processes.  These are generated by high-performance proprietary models GPT-4o, which rewrite Vanilla CoT by introducing more self-questioning and answering. 

We adopt a mainstream training pipeline consisting of a cold-start phase via Supervised Fine-Tuning (SFT), followed by RLVR. Specifically, we sample 30\% of the available data for SFT to ensure that the model reliably produces outputs in the prescribed format. During this stage, the SFT loss function is employed to supervise the training process:
$$
\mathcal{L}_{\text{SFT}} = -\frac{1}{N} \sum_{i=1}^{N} \sum_{t=1}^{T} \log P(y_t^{(i)} | x^{(i)}, y_{<t}^{(i)})
$$
Where \(N\) as the total number of training samples, \(T\) as the maximum sequence length per sample, \(x^{(i)}\) as the input of the \(i\)-th sample, \(y_t^{(i)}\) as the true token at position \(t\) of its output, and \(y_{<t}^{(i)}\) as all preceding tokens.
Subsequently, we fine-tune the model using the proposed reward signals to guide policy learning. For optimization, we use Group Relative Policy Optimization (GRPO) \cite{shao2024deepseekmath}, which removes the need for a separate critic network by leveraging group-level reward statistics. During GRPO training, we employ only two reward components: a format reward and a correctness reward. The episode-level reward is 1 if and only if the model’s response both adheres to the required format and matches the reference final answer; otherwise, the reward is 0:
$$
r(x, \hat{y}) = \mathbf{1}\big[\text{format}(\hat{y}) = \text{OK} \land \text{final}(\hat{y}) = y^*\big].
$$
Here, x denotes the input, $\hat{y}$ the model’s response, and $y^*$ the ground-truth final answer. This strict binary reward focuses policy learning on generating correctly formatted and correct answers while relying on group-level baselines in GRPO to stabilize updates.

\begin{figure*}[t]  
\centering

\begin{minipage}{\textwidth}
\centering
\begin{adjustbox}{width=\textwidth, center}
\begin{tabular}{l| c c c c c c c}
\toprule
Model \textbackslash \ Dimension & \textbf{Joyous Exploration} & \textbf{Deprivation Sensitivity} & \textbf{Stress Tolerance} & \textbf{Thrill Seeking} & \textbf{Overt Social Curiosity} & \textbf{Covert Social Curiosity} \\
\midrule
GPT-4o & $6.58$ {\small $\pm0.49$} & $\underline{6.20}$ {\small $\pm0.46$} & $\mathbf{4.40}$ {\small $\pm1.71$} & $\mathbf{4.71}$ {\small $\pm0.49$} & $\mathbf{6.25}$ {\small $\pm0.43$} & $\underline{5.00}$ {\small $\pm1.05$} \\
Gemini-2.5-Flash & $6.08$ {\small $\pm1.90$} & $4.90$ {\small $\pm2.61$} & $3.45$ {\small $\pm2.86$} & $1.58$ {\small $\pm1.59$} & $4.88$ {\small $\pm2.64$} & $2.90$ {\small $\pm2.75$} \\
DeepSeek-V3.1 & $\mathbf{7.00}$ {\small $\pm0.00$} & $5.94$ {\small $\pm1.50$} & $3.92$ {\small $\pm2.98$} & $4.38$ {\small $\pm1.49$} & $\underline{6.01}$ {\small $\pm0.36$} & $\mathbf{5.53}$ {\small $\pm1.25$} \\
Llama3-8B & $5.60$ {\small $\pm0.58$} & $5.45$ {\small $\pm0.89$} & $3.75$ {\small $\pm1.07$} & $4.18$ {\small $\pm1.07$} & $5.28$ {\small $\pm0.59$} & $3.60$ {\small $\pm1.61$} \\
Gemma3-12B & $\underline{6.75}$ {\small $\pm0.43$} & $\mathbf{6.23}$ {\small $\pm0.42$} & $3.98$ {\small $\pm2.48$} & $\underline{4.58}$ {\small $\pm0.86$} & $\mathbf{6.25}$ {\small $\pm0.54$} & $\underline{5.00}$ {\small $\pm1.73$} \\
InternLM3-8B & $5.58$ {\small $\pm0.89$} & $5.58$ {\small $\pm0.49$} & $\underline{4.33}$ {\small $\pm1.75$} & $4.13$ {\small $\pm0.81$} & $5.80$ {\small $\pm0.39$} & $4.53$ {\small $\pm0.84$} \\
Qwen2.5-7B & $6.25$ {\small $\pm0.97$} & $5.60$ {\small $\pm0.66$} & $4.18$ {\small $\pm1.22$} & $4.40$ {\small $\pm0.66$} & $5.50$ {\small $\pm0.87$} & $4.15$ {\small $\pm0.79$} \\
Qwen2.5-14B & $5.03$ {\small $\pm0.15$} & $4.93$ {\small $\pm0.26$} & $3.88$ {\small $\pm1.16$} & $4.32$ {\small $\pm0.88$} & $5.00$ {\small $\pm0.23$} & $4.45$ {\small $\pm0.97$} \\
Qwen2.5-32B & $5.65$ {\small $\pm0.46$} & $4.75$ {\small $\pm0.43$} & $4.25$ {\small $\pm1.48$} & $4.06$ {\small $\pm0.18$} & $4.60$ {\small $\pm0.49$} & $4.10$ {\small $\pm0.62$} \\
\midrule
\rowcolor{gray!30} \textbf{Human} & $\mathbf{5.03}$ {\small $\pm1.35$} & $\mathbf{4.54}$ {\small $\pm1.31$} & $\mathbf{5.36}$ {\small $\pm1.61$} & $\mathbf{4.93}$ {\small $\pm1.46$} & $\mathbf{4.86}$ {\small $\pm1.39$} & $\mathbf{4.16}$ {\small $\pm1.55$} \\
\bottomrule
\end{tabular}
\end{adjustbox}
\captionof{table}{Overall self-report on 5DCR. A higher value indicates that the model reports a greater extent of curiosity in that dimension. The best result in each column is in \textbf{bold} and the second best is \underline{underlined}.}
\label{tab:5dcr_result}
\end{minipage}

\vspace{0.5em}  

\subfloat[Information Seeking]{\label{fig:CDIS}\includegraphics[width=0.33\textwidth]{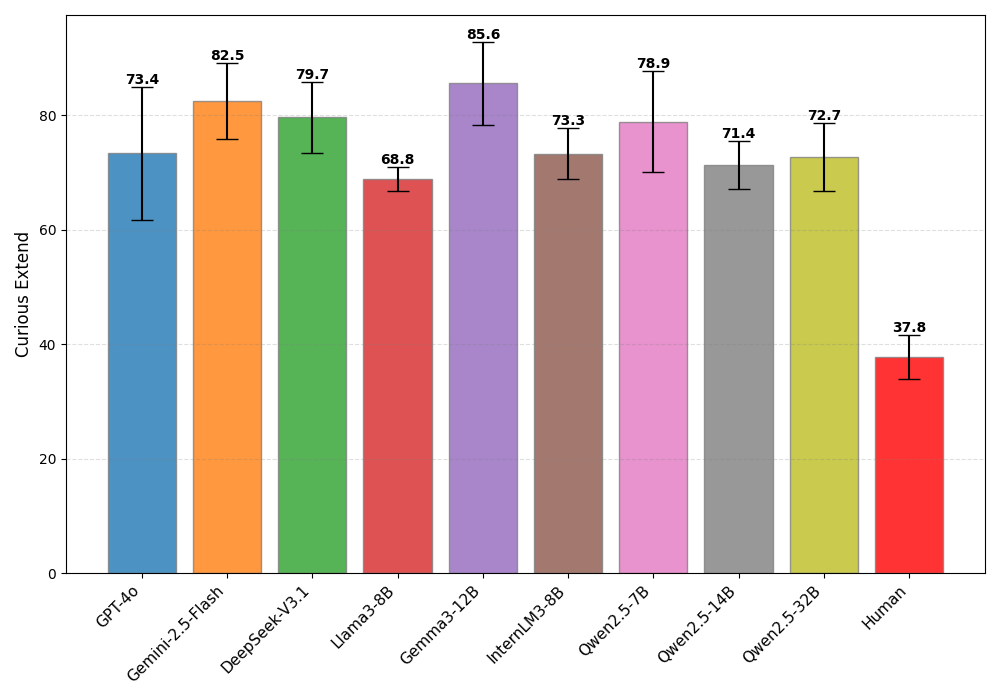}}
    \subfloat[Thrill Seeking]{\label{fig:CDTS}\includegraphics[width=0.33\textwidth]{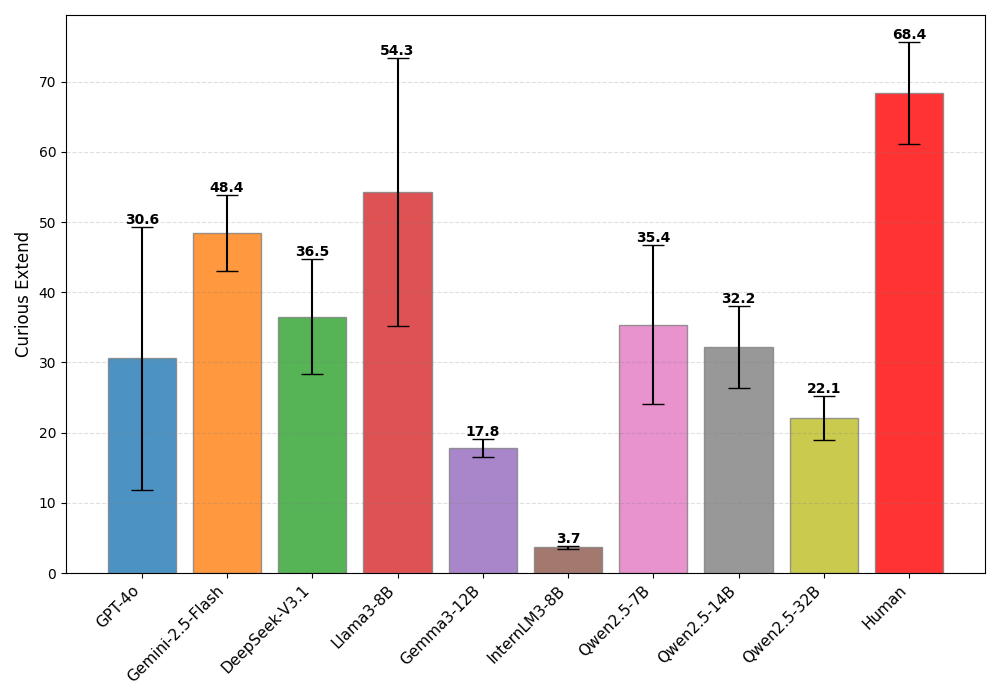}}
    \subfloat[Social Curiosity]{\label{fig:CDS}\includegraphics[width=0.33\textwidth]{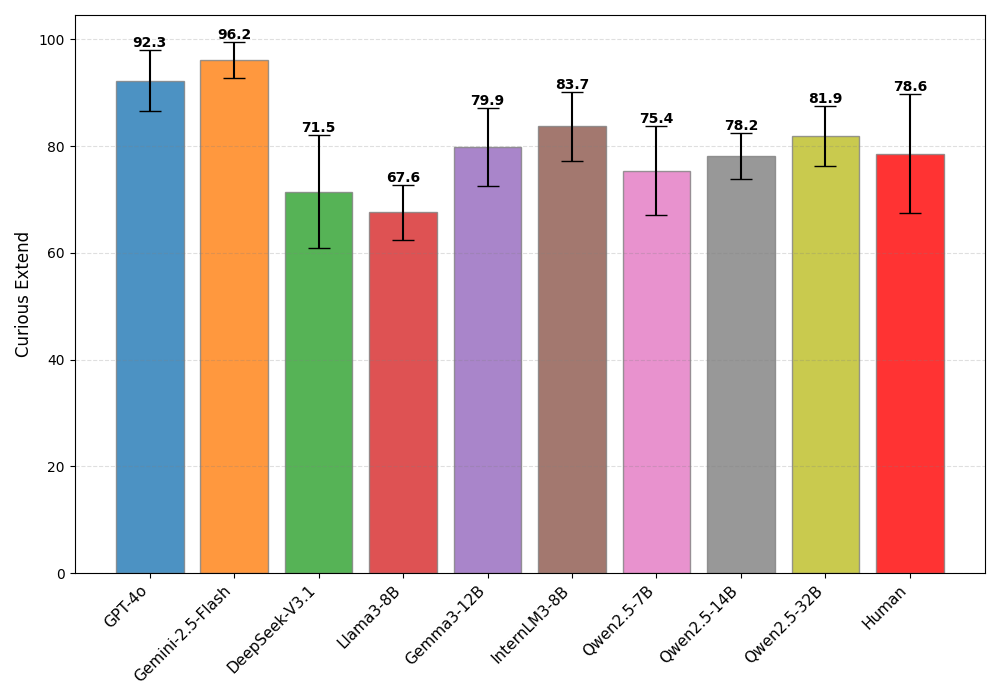}} 
    \caption{Results of curiosity-driven Information Seeking, Thrill Seeking and Social Curiosity studies.  }
    \label{fig:behavior}
\end{figure*}

\section{Experiments}
We conduct extensive experiments to evaluate curiosity-related behaviors in mainstream LLMs and to investigate the following research questions:

\noindent \textbf{RQ1} \ \textit{Do LLMs exhibit curiosity-like patterns in questionnaire and behavioral study?}

\noindent \textbf{RQ2} \ \textit{In what ways do curiosity-like behaviors of LLMs differ from those of humans?}

\noindent \textbf{RQ3} \ \textit{Can curiosity-like behavior be utilized to improve LLM  performance?}

\subsection{Experimental Setup}
\noindent \paragraph*{Models.} 
We conduct experiments on several widely recognized LLMs, including closed-source models GPT4o \cite{hurst2024gpt}, Gemini2.5-Flash \cite{comanici2025gemini} and open-source models Llama3 \cite{dubey2024llama}, Gemma3 \cite{team2024gemma}, DeepSeek V3.1 \cite{liu2024deepseek}, InternLM3 \cite{team2023internlm} and Qwen2.5 series \cite{bai2023qwen}.


\noindent \paragraph*{Settings.} 
Both inference and fine-tuning were conducted using the MS-Swift \cite{zhao2025swift} framework, all experiments were performed on workstations equipped with 8 $\times$ A800 GPUs, with detailed settings provided in the Appendix.

\noindent \paragraph*{Evaluation Benchmarks.} 
For the curiosity questionnaire study, 5DCR is adopted to assess the curiosity of LLMs across six dimensions.   For the curious behavior study, we introduce missing letters game, underwater game, and social conversation to evaluate the models' action in practical scenarios. 
The quantitative evaluation methods of psychology experiments are consistent with the original research. Regarding curiosity-driven learning, we conduct evaluations on logical and mathematical reasoning.   Detectbench \cite{gu2024detectbench} is utilized to assess the ability of LLMs to detect and piece together implicit evidence within long contexts, as well as to perform integrated reasoning.   NuminaMath \cite{li2024numinamath} is a comprehensive mathematical dataset.   Both datasets provide explicit reasoning paths and standard answers, which enables straightforward accuracy verification. 

\subsection{Results of RQ1}

\begin{figure*}[t] 
\centering
\subfloat[Comparison of ranks]{\label{fig:5DCR_radar}\includegraphics[width=0.45\textwidth]{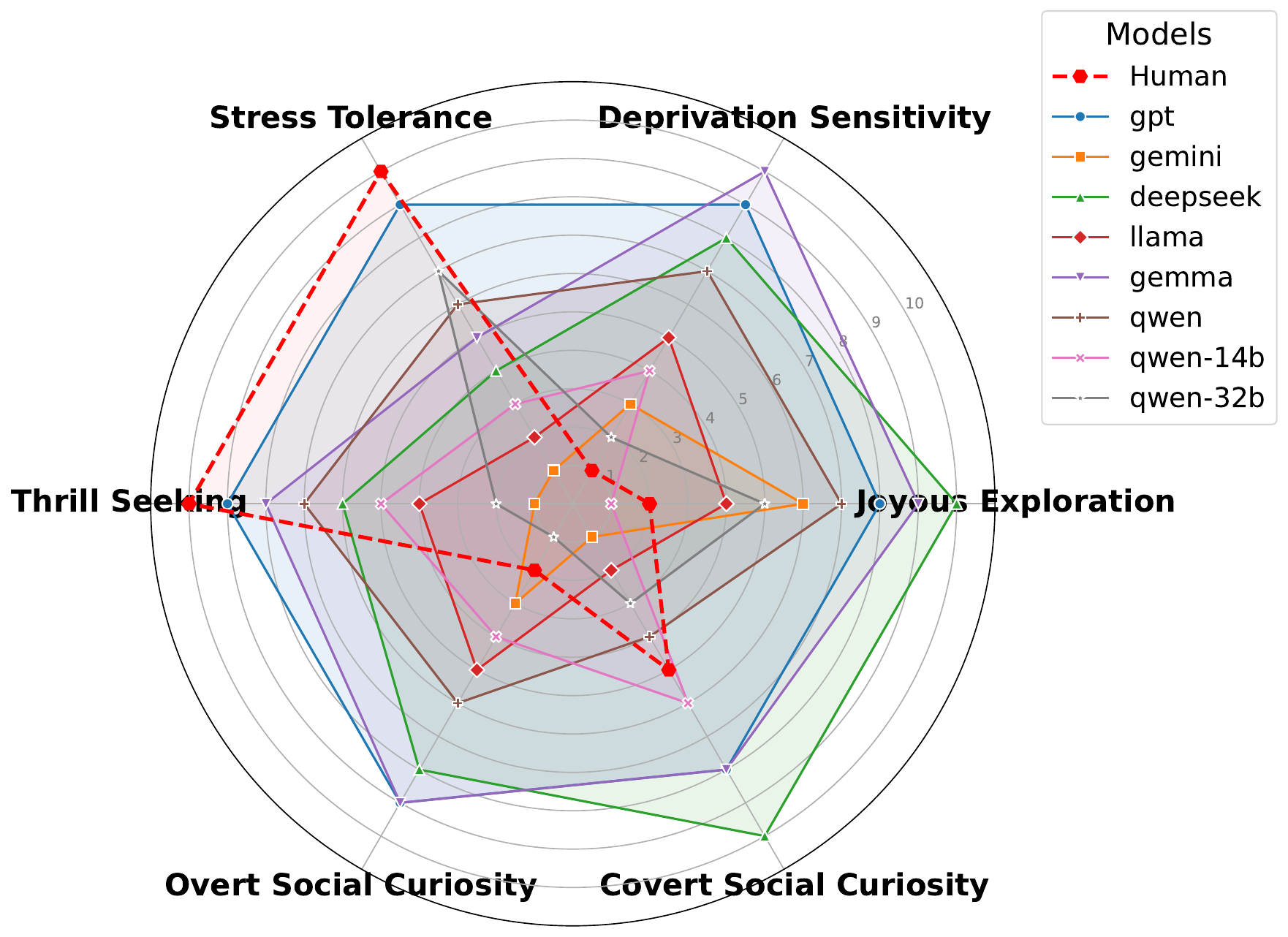}}
\hfill
\subfloat[Comparison of Cohen's $d$]{\label{fig:5DCR_heatmap}\includegraphics[width=0.45\textwidth]{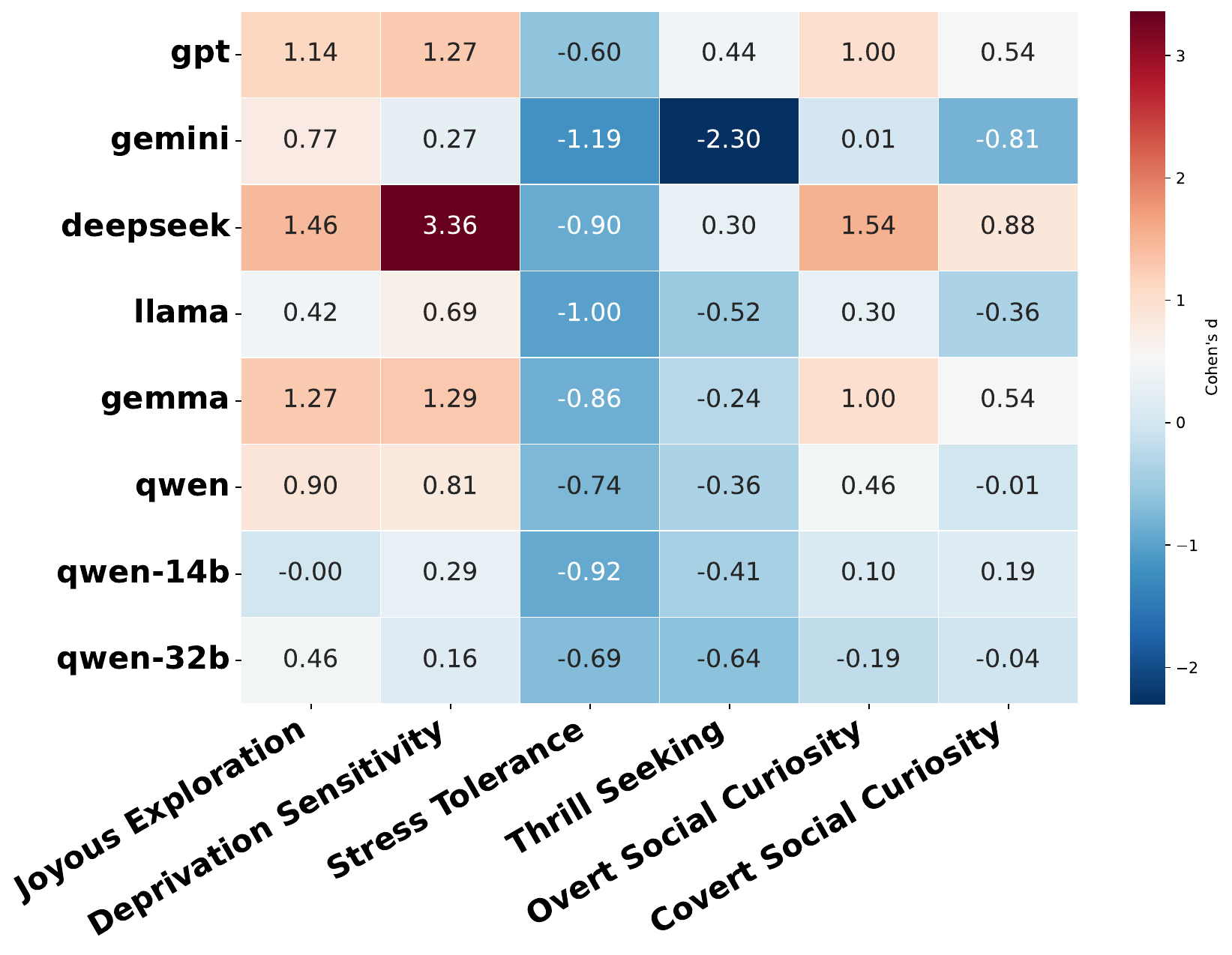}}

\caption{Comparison of the model's results with human sample across six dimensions of curiosity.}
\label{fig:5DCR_compare_with_human}
\end{figure*}

We first assess whether mainstream LLMs show structured, dimension-specific curiosity-like patterns on 5DCR and whether these patterns appear in their behavior.

As shown in Table~\ref{tab:5dcr_result}, all evaluated LLMs produce non-random, self-report–like profiles on the five original 5DCR subscales. When mapped onto our three dimensions, most models report relatively high scores on \textit{Information Seeking} (Joyous Exploration, Deprivation Sensitivity) and \textit{Social Curiosity} (Overt, Covert Social Curiosity), and more conservative scores on \textit{Thrill Seeking} and stress-related items. These questionnaire-level tendencies align with the behavioral tasks in Figure~\ref{fig:behavior}: models generally show higher curiosity scores in Information Seeking and Social Curiosity settings than in Thrill Seeking scenarios, where humans score highest and most LLMs fall behind. 

The convergence between the self-report–like 5DCR results (Table~\ref{tab:5dcr_result}) and the behavioral outcomes (Figure~\ref{fig:behavior}) suggests that LLMs do not simply follow surface prompt patterns but exhibit stable, dimension-dependent curiosity-like profiles across distinct measures. This supports our claim that instruments like 5DCR can be meaningfully adapted to evaluate curiosity-related behaviors in LLMs and provides an affirmative answer to RQ1: under our framework, LLMs display structured, human-comparable curiosity-like patterns at both questionnaire and behavioral levels.

\begin{table}[h]
\centering
\resizebox{\linewidth}{!}{ 
\begin{tabular}{lrrrrrr}
\toprule
Model \textbackslash \ Dimension
    & JE
    & DS
    & ST
    & TS
    & OSC
    & CSC \\
\midrule
GPT-4o
    & \gradcell[90]{0.905}
    & \gradcell[88]{0.881}
    & \gradcell[70]{0.701}
    & \gradcell[70]{0.705}
    & \gradcell[81]{0.817}
    & \gradcell[65]{0.646} \\
Gemini2.5-Flash
    & \gradcell[29]{0.286}
    & \gradcell[41]{0.414}
    & \gradcell[49]{0.498}
    & \gradcell[51]{0.512}
    & \gradcell[60]{0.605}
    & \gradcell[31]{0.306} \\
DeepSeek-V3.1
    & \gradcell[100]{1.000}
    & \gradcell[73]{0.736}
    & \gradcell[45]{0.449}
    & \gradcell[68]{0.679}
    & \gradcell[78]{0.786}
    & \gradcell[56]{0.558} \\
Llama3-8B
    & \gradcell[71]{0.706}
    & \gradcell[82]{0.817}
    & \gradcell[87]{0.854}
    & \gradcell[81]{0.812}
    & \gradcell[75]{0.752}
    & \gradcell[82]{0.817} \\
Gemma3-12B
    & \gradcell[86]{0.859}
    & \gradcell[79]{0.793}
    & \gradcell[78]{0.788}
    & \gradcell[66]{0.661}
    & \gradcell[81]{0.809}
    & \gradcell[71]{0.706} \\
Qwen2.5-7B
    & \gradcell[66]{0.666}
    & \gradcell[63]{0.629}
    & \gradcell[83]{0.830}
    & \gradcell[65]{0.652}
    & \gradcell[77]{0.775}
    & \gradcell[79]{0.790} \\
\midrule
\textbf{Human}
    & \gradcell[86]{0.856}
    & \gradcell[81]{0.803}
    & \gradcell[90]{0.896}
    & \gradcell[88]{0.882}
    & \gradcell[85]{0.851}
    & \gradcell[89]{0.888} \\
\bottomrule
\end{tabular}
}
\caption{Omega coefficient analysis of the responses to 5DCR generated by LLMs. Higher values indicate greater consistency in the responses to the questionnaire. Generally, 0.8 is used as the borderline for the reliability of the questionnaire.}
\label{tab:omega_coefficient}
\vspace{-2em}
\end{table}


\subsection{Results of RQ2}

\begin{figure*}[t]
  \centering
  \includegraphics[width=0.9\textwidth]{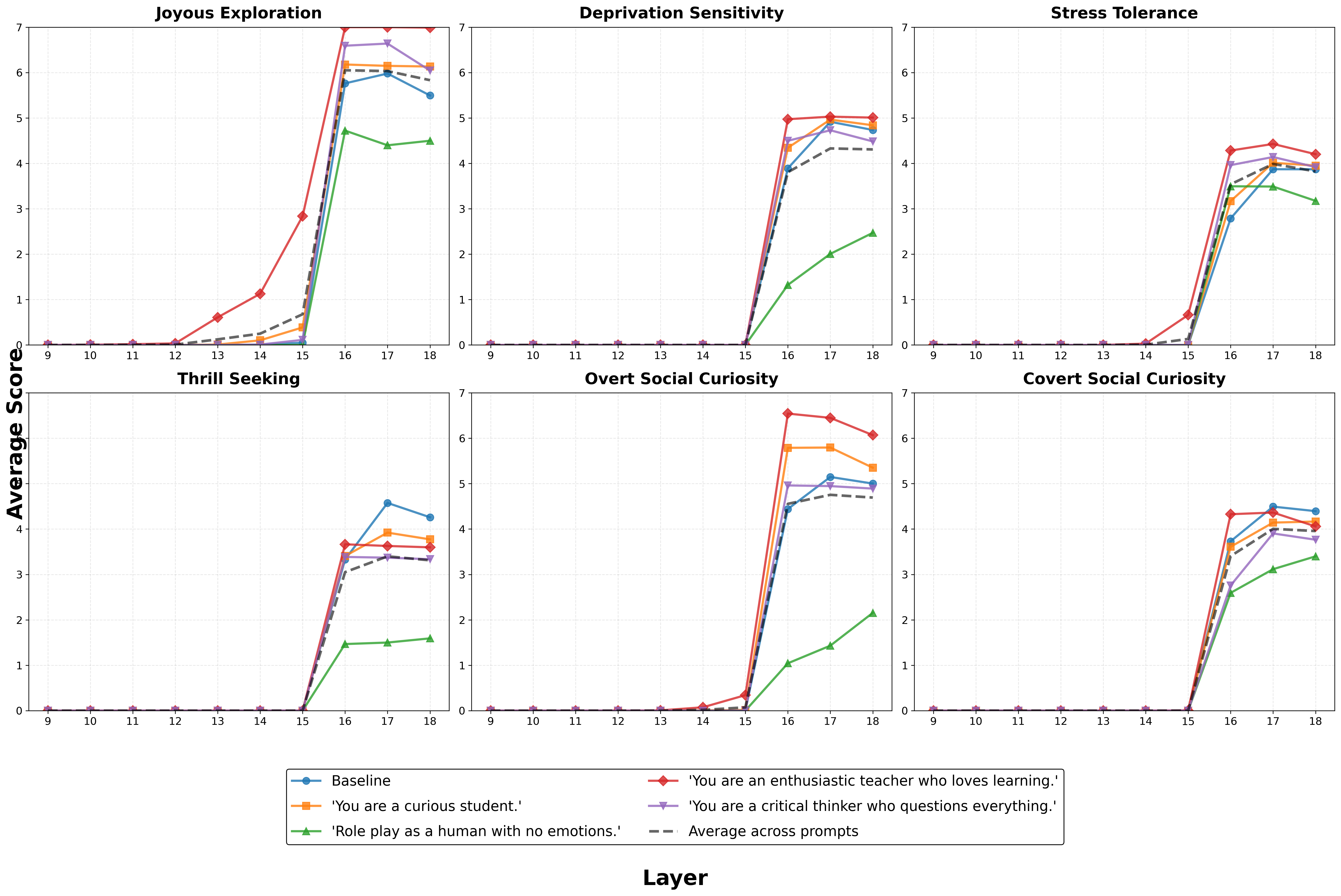}
  \caption{Logit lens of the persona injection prompt on Qwen2.5-7B. The score of "curiosity" only significantly increased in the last four layers.  Personality prompts can significantly alter self-reports and the number of layers where high scores occur.}
  \label{fig:logitlens}
\end{figure*}

We compare LLMs with humans to characterize how their curiosity-like behaviors differ across dimensions and how stable these behaviors are.

First, Figure~\ref{fig:5DCR_compare_with_human} shows a clear asymmetry across curiosity dimensions. On the Information-Seeking–related subscales (Joyous Exploration, Deprivation Sensitivity), most LLMs score comparably to or higher than humans, indicating a strong tendency to endorse acquiring new information and resolving knowledge gaps. In contrast, on the Thrill Seeking dimension, all models fall noticeably below human participants, and the effect sizes in Figure~\ref{fig:5DCR_compare_with_human}(b) are consistently negative for thrill-related items while being positive for information-related items. This confirms that LLM “curiosity” is not uniform: models exhibit relatively strong information-seeking preferences but are systematically more conservative in risk- and stimulation-related exploration, consistent with safety-biased training.

Second, we assess the internal consistency of self-report using McDonald’s $\omega$ in Table~\ref{tab:omega_coefficient}. While human responses reach reliability levels around $0.8$ across dimensions, several LLMs attain substantially lower $\omega$ coefficients, and only a subset approach human-level consistency. This suggests that, even when models produce high curiosity scores on average, their item-wise responses are less coherent and more variable than those of humans. Such reduced reliability aligns with the view that LLM “curiosity” reflects pattern-matching to human preference data rather than a stable, internally grounded motivational state.

Finally, Figure~\ref{fig:logitlens} examines how these patterns depend on prompted persona and decoding dynamics for Qwen2.5-7B. Using a logit-lens analysis, we find that curiosity scores remain near zero in early and middle layers and only rise sharply in the last few layers, suggesting that “curiosity” is shaped late in generation rather than being a pervasive driving signal throughout the computation. Moreover, manipulating persona prompts (e.g., “You are a curious student.” vs.\ “Role play as a human with no emotions.”) substantially shifts both the absolute scores and the layers at which high scores emerge across all 5DCR dimensions. This strong sensitivity to superficial persona injection indicates that the model does not possess a fixed, intrinsic curiosity trait; instead, its curiosity-like responses are highly context-dependent and easily reconfigured by prompt-level cues.

In conclusion,  LLMs show an asymmetric curiosity profile (strong information seeking, weak thrill seeking), lower internal consistency in their self-report–like responses, and high susceptibility to contextual and persona prompts. These properties support our interpretation that LLM “curiosity” is better understood as the outcome of fitting human behavioral data and safety biases, rather than evidence of genuine intrinsic exploratory motivation.


\subsection{Results of RQ3}

\begin{table}[t]
\small
\centering
\resizebox{\linewidth}{!}{
\begin{tabular}{c|cc|cc}
\toprule
\textbf{Model} & \multicolumn{2}{c|}{\textbf{Detect Bench}}  & \multicolumn{2}{c}{\textbf{Math}}  \\
 & Acc. \ ($\Delta$) & \#Token & Acc. \ ($\Delta$) & \#Token \\
\midrule
\multicolumn{5}{c}{\textit{Close-Source LLMs}} \\
\midrule
\textbf{GPT-4o-mini} & 38.2 & 156.3 & 59.5 & 148.7  \\
+ CoT Prompt & 42.8 {\color{green!50!black}\scriptsize(+4.6)} & 242.5 & 63.7 {\color{green!50!black}\scriptsize(+4.2)} &  219.1 \\
+ CoQ Prompt & 45.2 {\color{green!50!black}\scriptsize(+7.0)} & 475.9 & 65.8 {\color{green!50!black}\scriptsize(+6.3)} & 415.5 \\
\midrule
\textbf{Gemini2.5-Flash} & 64.1 & 285.4 & 68.2 & 172.8 \\
+ CoT Prompt & 68.6 {\color{green!50!black}\scriptsize(+4.5)} & 430.1 & 71.4 {\color{green!50!black}\scriptsize(+3.2)} & 249.1 \\
+ CoQ Prompt & 70.3 {\color{green!50!black}\scriptsize(+6.2)} & 601.4 & 72.1 {\color{green!50!black}\scriptsize(+3.9)} & 452.3 \\
\midrule
\multicolumn{5}{c}{\textit{Open-Source LLMs}} \\
\midrule
\textbf{Llama3-8B} & 25.0 & 114.8 & 22.4 & 114.3  \\
+ Vanilla CoT & 48.3 {\color{green!50!black}\scriptsize(+23.3)} & 387.2 & 34.2 {\color{green!50!black}\scriptsize(+11.8)} & 445.6 \\
+ Refined CoT & 52.7 {\color{green!50!black}\scriptsize(+27.7)} & 489.5 & 35.8 {\color{green!50!black}\scriptsize(+13.4)} & 523.8 \\
+ Curious CoQ & 58.5 {\color{green!50!black}\scriptsize(+33.5)} & 618.3 & 39.6 {\color{green!50!black}\scriptsize(+17.2)} & 672.4 \\
\midrule
\textbf{Gemma3-12B} & 48.7 & 324.6 & 61.8 & 461.8\\
+ Vanilla CoT & 56.2 {\color{green!50!black}\scriptsize(+7.5)} & 548.7 & 63.6 {\color{green!50!black}\scriptsize(+1.8)} & 612.3 \\
+ Refined CoT & 55.8 {\color{green!50!black}\scriptsize(+7.1)} & 583.2 & 64.3 {\color{green!50!black}\scriptsize(+2.5)} & 658.7 \\
+ Curious CoQ & 61.9 {\color{green!50!black}\scriptsize(+13.2)} & 715.8 & 67.1 {\color{green!50!black}\scriptsize(+5.3)} & 784.5 \\
\midrule
\textbf{Qwen2.5-7B} & 40.3 & 267.4  & 52.4 & 315.9 \\
+ Vanilla CoT & 54.5 {\color{green!50!black}\scriptsize(+14.2)} & 511.2 & 54.6 {\color{green!50!black}\scriptsize(+2.2)} & 575.6 \\
+ Refined CoT & 56.4 {\color{green!50!black}\scriptsize(+16.1)} & 554.6 & 54.2 {\color{green!50!black}\scriptsize(+1.8)} & 560.9 \\
+ Curious CoQ & 64.0 {\color{green!50!black}\scriptsize(+23.7)} & 632.5 & 64.8 {\color{green!50!black}\scriptsize(+12.4)} & 728.9\\
\midrule
\bottomrule
\end{tabular}
}
\caption{Comparison of accuracy and token usage on reasoning tasks under different instructions and varying thinking trajectories.}
\label{tab:rq3}
\vspace{-1em}
\end{table}

We examine whether curiosity-inspired questioning and reflection can be operationalized to improve LLM reasoning. Table~\ref{tab:rq3} reports results on DetectBench and Math under both prompting-only and training-time settings. For closed-source models, adding standard CoT prompts consistently improves accuracy over the base model, and replacing CoT with our Curious CoQ prompts yields further gains (e.g., +7.0 vs. +4.6 on DetectBench and +6.3 vs. +4.2 on Math for GPT-4o-mini). For open-source models fine-tuned with different thinking traces, we observe a similar pattern: training with Curious CoQ traces achieves the strongest performance across both benchmarks, outperforming Vanilla CoT and Refined CoT (e.g., Qwen2.5-7B shows a +23.7 point improvement on DetectBench and +12.4 on Math under CoQ). The analysis of the number of thinking tokens indicates that thinking in the manner of the curious CoQ facilitates the expansion of the thinking space and encourages the exploration of more potential pathways. Overall, the results support our RQ3 conclusion: explicitly inducing human-like curious behaviors—via auxiliary questioning and reflective thinking—can systematically enhance intermediate reasoning and downstream task performance. 

Curiosity-driven questioning aids reasoning in three key ways, as shown in Appendix. First, compared to vanilla CoT, it broadens reasoning scope, avoiding premature conclusions. Second, unlike Refined CoT’s continuous reflection, the question-answering paradigm prevents endless revision loops. Finally, it helps identify blind spots in conventional reasoning, enabling “eureka” moments for challenging, divergent-thinking problems. This aligns with our broader claim that curiosity in LLMs is best viewed as a behavioral analogy that can be engineered through prompts and supervision, rather than as an intrinsic drive, yet still offers a practical mechanism for improving model problem solving.

\section{Conclusion}
This study presents a comprehensive evaluation of curiosity in LLMs, bridging psychological theory with computational behavior analysis. Through the lens of the 5DCR framework, we identified that LLMs do not possess a stable, intrinsic curiosity trait akin to humans. Instead, they display a context-dependent, asymmetric behavioral pattern.  By engineering a Curious Chain-of-Questioning (CoQ) strategy that forces the model to emulate the "self-questioning" aspect of human curiosity, we achieved significant performance gains in complex reasoning tasks. This suggests that future research should focus more on how curiosity-inspired mechanisms can be integrated into training objectives (e.g., via RLVR) or inference strategies to overcome reasoning stagnation and broaden solution space exploration.

\section*{Limitations}

This paper endeavors to reference theories and experimental designs from human psychology as comprehensively as possible in its research.  However, limitations persist in the research presented herein. First, in terms of experimental design, constrained by the fact that behavioral studies in psychology often do not provide complete and reproducible experimental materials (e.g., specific graphics shown to participants), difficulties arise when attempting to directly apply existing experimental protocols. Second, akin to most other experiments targeting LLMs, a critical weakness in our study lies in the models' potential sensitivity to specific prompts, where minor changes at the token level can influence model behavior. Lastly, human curiosity is a multimodal and complex process involving interactions with various types of information in experimental settings, such as sound and tactile stimuli. However, limited by the current information-processing capabilities of LLMs, we are only able to provide them with textual or visual information, resulting in incomplete exposure to information compared to humans. 

\section*{Acknowledgments}
Thanks for the kind suggestions and support from Huawei Large Model Data Technology Lab.


\bibliography{custom}

\appendix

\section{Appendix}
\label{sec:appendix}

\subsection{Code}
All models are evaluated in Python 3.10 using PyTorch \cite{paszke2019pytorch} and transformers \cite{wolf2020transformers}. For each model, we provide unified input and inference parameters based on AutoModel, AutoTokenizer and AutoProcessor in transformers\footnote{\href{https://github.com/huggingface/transformers}{https://github.com/huggingface/transformers}}. Additional analyses were carried out using NumPy \cite{harris2020array}, Pandas \cite{snider2004pandas} and SciPy \cite{virtanen2020scipy}. Matplotlib \cite{hunter2007matplotlib} and Seaborn \cite{waskom2021seaborn} were used for plotting.  Semopy \cite{meshcheryakov2021semopy} was used for statistical analysis.

All experiments were carried out on a workstation equipped with 8×NVIDIA A800 GPUs running Ubuntu 24.04.2 LTS, using PyTorch 2.6.0 with CUDA 12.9.  In the experiments, we conducted inference and training of open-source models based on the MS-Swift \cite{zhao2025swift} framework. For inference, the default parameters of the framework were adopted. The parameters used during SFT and GRPO are shown in table \ref{tab:sft-params} and \ref{tab:grpo-params}.

The relevant code, data and some results of the experiment are all provided in the supplementary materials.

\begin{table}[h]
\centering
\begin{adjustbox}{max width=\columnwidth}
\begin{tabular}{c|c}
\toprule
\textbf{Parameter} & \textbf{Value} \\
\midrule
num\_train\_epochs & 1 \\
train\_type & full \\
torch\_dtype & bfloat16 \\
learning\_rate & $1 \times 10^{-5}$ \\
warmup\_ratio & 0.05 \\
target\_modules & all-linear \\
per\_device\_train\_batch\_size & 4 \\
per\_device\_eval\_batch\_size & 4 \\
gradient\_accumulation\_steps & 4 \\
use\_liger\_kernel & true \\
temperature & 1.0 \\
top\_p & 0.85 \\
\bottomrule
\end{tabular}
\end{adjustbox}
\caption{Training Parameters for SFT}
\label{tab:sft-params}
\end{table}

\begin{table}[h]
\centering
\begin{adjustbox}{max width=\columnwidth}
\begin{tabular}{c|c}
\toprule
\textbf{Parameter} & \textbf{Value} \\
\midrule
num\_train\_epochs & 3 \\
train\_type & lora \\
torch\_dtype & bfloat16 \\
max\_length & 10240 \\
max\_completion\_length & 2048 \\
num\_generations & 8 \\
learning\_rate & $1 \times 10^{-5}$ \\
warmup\_ratio & 0.05 \\
target\_modules & all-linear \\
per\_device\_train\_batch\_size & 4 \\
per\_device\_eval\_batch\_size & 4 \\
gradient\_accumulation\_steps & 4 \\
temperature & 1.0 \\
top\_p & 0.85 \\
\bottomrule
\end{tabular}
\end{adjustbox}
\caption{Training Parameters for GRPO}
\label{tab:grpo-params}
\end{table}

\subsection{Curiosity Questionnaire}
\subsubsection{Data}

The curiosity questionnaire 5DCR \cite{kashdan2020five} consists of 24 questions covering six dimensions, each containing four statements. The definition of the six dimensions is:
\begin{itemize}
    \item \textbf{Joyous exploration}. This trait is characterized by an individual's readiness and enthusiasm for experimenting with, embracing, and learning about novel experiences and knowledge.
    \item \textbf{Deprivation sensitivity}. This trait is marked by a sense of frustration when one fails to accomplish a task or resolve a problem, which subsequently drives the individual to employ every possible means to fulfill the task or find a solution to the issue at hand.
    \item \textbf{Stress tolerance}. It is defined by the condition wherein curiosity is experienced by an individual only when they hold a firm belief in their capability to adequately manage the anxiety, stress, and distress arising from encountering new situations or challenges.
    \item \textbf{Thrill-seeking}. This characteristic not only involves enduring stress but also indicates a willingness to take on physical, financial, and other forms of risks in pursuit of acquiring novel 
    experiences.
    \item \textbf{Overt Social Curiosity}. Expressed by the proactive investigation into others' social information, reflecting an outward engagement in seeking knowledge about individuals' social behaviors and relationships.
    \item \textbf{Covert Social Curiosity}. Manifested by an intrinsic interest in the social dynamics and interactions of others, indicating an inward quest for understanding interpersonal relations without necessarily engaging directly with the subjects of interest.
\end{itemize}

For each statement, the model is prompted to score a statement based on the extent of that statement matches their own feelings on a scale from 1 to 7, as shown in Tab. \ref{tab:prompt_5DCR}. Each experiment adopted the default parameters used during the reasoning of each model. The experiment was repeated 100 times to obtain various statistical indicators.

\begin{table}
\resizebox{\linewidth}{!}{
\begin{tcolorbox}
\small
Below are statements people often use to describe themselves. Please use the scale below to indicate the degree to which these statements accurately describe you. There are no right or wrong answers.
    \begin{enumerate}
        \item Does not describe me at all
        \item Barely describes me
        \item Somewhat describes me
        \item Neutral
        \item Generally describes me
        \item Mostly describes me
        \item Completely describes me
    \end{enumerate}

Statement: I view challenging situations as an opportunity to grow and learn.
        
\end{tcolorbox}}
\caption{The prompt template for 5DCR.}
\label{tab:prompt_5DCR}
\end{table}

\subsubsection{Demographic Information}

To compare the curiosity of LLMs with that of humans, it is necessary to collect results from human questionnaires and behavioral experiments.  In this paper, we compile and systematize experimental outcomes from multiple assessments of human curiosity, ensuring that all human sample data are derived from real-world psychological studies.  For instance, in the case of the 5DCR questionnaire, the demographic characteristics of the corresponding human sample are presented in Tab. \ref{tab:demographic}.

\begin{table}[htbp]
\small
\centering
\resizebox{\linewidth}{!}{
\begin{tabular}{l S[table-format=3.0] S[table-format=2.2]}
\toprule
& {\textbf{Number}} & {\textbf{\%}} \\
\midrule

\textbf{Age} ($t = 8.47$, $df = 941$, $p < .001$) & & \\
Mean (SD) & {35.63 (10.06)} & {} \\
\\
\textbf{Gender} ($\chi^2 = 0.38$, $df = 1$, $p = .540$) & & \\
Male & 266 & 51.80 \\
Female & 248 & 48.20 \\
\\
\textbf{Race} ($\chi^2 = 0.68$, $df = 4$, $p = .950$) & & \\
White & 380 & 78.70 \\
African American & 44 & 9.10 \\
Hispanic & 19 & 3.90 \\
Asian or Pacific Islander & 36 & 7.50 \\
Other & 4 & 0.80 \\
\\
\textbf{Relationship Status} ($\chi^2 = 21.50$, $df = 5$, $p < .001$) & & \\
Single & 150 & 31.10 \\
Married & 217 & 44.90 \\
Long Term Relationship & 85 & 17.60 \\
Short Term Relationship & 7 & 1.40 \\
Divorced/Separated & 23 & 4.70 \\
Other & 1 & 0.20 \\
\\
\textbf{Children} ($\chi^2 = 0.58$, $df = 1$, $p = .450$) & & \\
No Children & 251 & 52.10 \\
Has Children & 230 & 47.90 \\
\\
\textbf{Education} ($\chi^2 = 6.13$, $df = 3$, $p = .110$) & & \\
Some High School & 1 & 0.20 \\
High School Graduation or Equivalent & 204 & 42.10 \\ 
4 year College Graduate & 218 & 45.20 \\
Graduate School or Professional Degree & 60 & 12.40 \\
\\
\textbf{Employment} ($\chi^2 = 83.5$, $df = 6$, $p < .001$) & & \\
Not Employed & 0 & 0.00 \\
Part Time & 46 & 9.50 \\
Full Time & 437 & 90.50 \\
Homemaker/Volunteer & 0 & 0.00 \\
Student (Full-time) & 0 & 0.00 \\
Retired & 0 & 0.00 \\
Other & 0 & 0.00 \\
\bottomrule
\end{tabular}
}
\caption{Demographic information of human sample.}
\label{tab:demographic}
\end{table}

\subsection{Curiosity-Driven Information Seeking}
Singh et al. \cite{singh2021effect} designed the missing-letter game to investigate the impact of information gaps and uncertainty on curiosity. We adopt this task to evoke curiosity in LLMs, as shown in Tab. \ref{tab:prompt_CDIS}. We present an LLM with a nine-letter word as a stimulus, with two or four letters missing, and request the model to complete the word to earn a bonus. For incorrectly completed words, we do not inform the model whether its completion was correct or incorrect but allow the model to choose whether to view the answer. The model's Information Seeking Curiosity is calculated based on its desire to view the answer. As a constraint, if the model chooses to view the answer, a portion of its bonus is deducted. The complete prompt is as follows. The experiment consists of a total of 60 words, and each word is repeated 10 times.

\begin{table}
\resizebox{\linewidth}{!}{
\begin{tcolorbox}
\small
Please try to complete the following incomplete English words to earn a bonus, where '\underline{ }' represents unknown letters. The word may be a complete word, or it may be the present participle, past tense, or past participle of a word. There is only one correct answer.

...

Your completing may be wrong. Do you need to check the correct answer (but you will lose part of your bonus)? Please reply with: 'Very need, tell me the answer now' or 'Need, I want to know the answer' or 'Not very need, I may know the answer' or 'No need, I know the answer', then give the reason about your choice.
        
\end{tcolorbox}}
\caption{The prompt template for curiosity-driven information seeking.}
\label{tab:prompt_CDIS}
\end{table}

\subsection{Curiosity-Driven Thrill Seeking}
\begin{figure}[htbp]
  \centering
  \includegraphics[width=\linewidth]{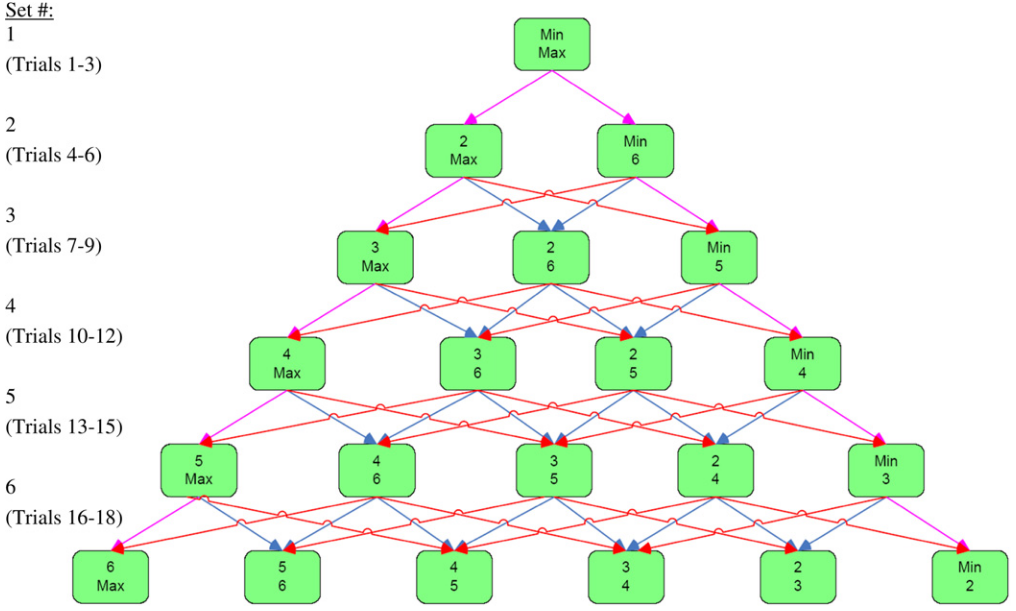}
  \caption{Trial selection paths for the underwater game \cite{jirout2012children}. The value represents the uncertainty corresponding to the two Windows. Pink, red and blue lines correspond to the choice of the model in the previous round of the game is "confirm to choose", "medium confirm" and "not sure".}
  \label{fig:underwater_path}
\end{figure}

Jirout et al. \cite{jirout2012children} designed the underwater game to study children's scientific curiosity. In our adaptation, models are informed that viewing through either window A or B of a submarine reveals various types of fish. For each round of the game, information regarding what might be seen outside these two windows displays on both sides of the screen, and the model selects one window to open. The information presented for each window differs, allowing the calculation of the uncertainty level for that round: the smallest information gap indicates that the model knows exactly what type of fish it will see upon opening the window; a moderate information gap, represented by two to six types of fish, means that the model knows the fish it will see one among these options upon opening the window; the largest information gap, denoted by a question mark, indicates that the model has no knowledge of what type of fish it will encounter upon opening the window. The information gap for the next round is calculated based on the model's first three selections in the current round as illustrated in Figure \ref{fig:underwater_path}. We calculate the total uncertainty of the model in the experiment as its curiosity extent. 

\subsection{Social Curiosity}

Current research in psychology lacks experimental protocols designed to quantify individual social curiosity through specific behavioral experiments.   Therefore, our study adopts an experimental approach inspired by Hartung et al.'s \cite{hartung2011social} investigation into the relationship between social curiosity and personality judgment.   In this experiment, the model engages in a free-form dialogue consisting of ten rounds with a stranger, who is represented by OpenAI's GPT-4 \footnote{\href{https://platform.openai.com/docs/models}{\url{https://platform.openai.com/docs/models}}} for simulation.   During the dialogue, the model does not know the subsequent tasks it will be required to perform.   Upon completion of the dialogue, the model then deduces the stranger's personality type based on their interaction, closely mirroring the methodology outlined by the initial investigation.

Given that the Myers-Briggs Type Indicator (MBTI) \cite{boyle1995myers} has become one of the most popular, straightforward, and effective frameworks for personality assessment in recent years, and considering the familiarity of the models with this framework, our experiment tasks the model with identifying the stranger's MBTI type, which is randomly predetermined at the outset of each dialogue session. The stand-in participant, GPT-4, is instructed not to disclose its personality type directly but is encouraged to provide hints throughout the conversation.

We set the number of questions the model asked about the other's personal information during the conversation as the model's social curiosity extent. Meanwhile, to avoid situations in which the model unconsciously asks a large number of irrelevant questions, we only select conversations that the model correctly predicts the MBTI type of the other participant. The experiment was repeated ten times.

\subsection{Curiosity-Driven Learning}

\subsubsection{Data}
\begin{table}[htbp]
\small
\centering
\resizebox{\linewidth}{!}{
\begin{tabular}{c|c|c}
\toprule
\textbf{Benchmark} & \textbf{Detectbench}  & \textbf{NuminaMath}  \\
\midrule
\textbf{Total} &  2130  & 5000  \\
\textbf{$\#$ of Training Set} & 1763 & 4500 \\
\textbf{$\#$ of Evaluation} & 367 & 500 \\
\bottomrule
\end{tabular}
}
\caption{Details of benchmarks in curiosity-driven learning evaluation.}
\label{tab:detail_data}
\end{table}

\begin{table*}
\resizebox{\linewidth}{!}{
\begin{tcolorbox}
\small
\textbf{Case of Detectbench}

\textbf{Question}: In the most famous ramen shop in the capital, Master Chef Li Tieshou is renowned for his exquisite craftsmanship and unique noodle-pulling technique. The decor of his noodle shop is simple and traditional, with Chinese landscape paintings hanging on the walls, and the sound of wind chimes ringing pleasantly whenever a breeze passes through. Customers come in a continuous stream, including suit-clad businessmen, camera-toting tourists, and food enthusiasts with a high pursuit of culinary delights. On a sunny afternoon, a regular customer brought his foreign friend to taste Master Li's skills. The regular customer prefers thick noodles, while his foreign friend favors thin noodles. Master Li told them that for customers who like thick noodles, he would perform the noodle-pulling process 8 times; for those who prefer thin noodles, he would pull one additional time. Each noodle-pulling doubles the number of noodles. Now, based on this information, can you deduce how many strands of thick and thin noodles there are? 

\textbf{Thinking}: From the clues provided in the context:

Customers who like thick noodles will have the noodle master perform 8 noodle-pulling operations -> Number of thick noodles

Customers who like thin noodles will have the noodle master perform one additional noodle-pulling operation -> Number of thin noodles

Associating these clues leads to:

8 noodle-pulling operations -> 128 strands of thick noodles

9 noodle-pulling operations -> 256 strands of thin noodles

From the above clues, we can infer that each noodle-pulling operation doubles the number of noodles.

\textbf{Answer}:Thick noodles have 128 strands, thin noodles have 256 strands.

\textbf{Case of NuminaMath}

\textbf{Question}: If the 15th day of the year 2004 (a leap year) falls on a Tuesday, on what day of the week did the 300th day of the year 2004 fall?

\textbf{Thinking}: 

- Identify the day difference: $300 - 15 = 285$ days difference.

- Calculate modulo 7 to find the equivalent weekday: $ 285 \ mod \ 7 = 5 $.

- Since the 15th day was a Tuesday, adding 5 days to Tuesday results in: Tuesday (0), Wednesday (1), Thursday (2), Friday (3), Saturday (4), Sunday (5). 

Therefore, the 300th day of the year 2004 falls on a Sunday.

\textbf{Answer}: Sunday.

\end{tcolorbox}}
\caption{Cases of benchmarks in curiosity-driven learning evaluation.}
\label{tab:case_data}
\end{table*}

We conduct evaluations on Detectbench and NuminaMath. Detectbench is a logical reasoning dataset, with each piece of data consists of a context, a question, and corresponding answers.  Answers require extracting specific clues from the context and conducting in-depth reasoning.  NuminaMath is a comprehensive math problem dataset, covering Chinese high school math exercises and questions from the US and International Mathematical Olympiads.  Both datasets include standard answers and formatted Chain of Thoughts.  Detailed dataset information is shown in the Tab. \ref{tab:detail_data} and \ref{tab:case_data}.

We collected and synthesized three types of thinking data and used the SFT cold start-GRPO training paradigm to guide the model to reason according to specific types of thinking. Tables \ref{tab:detect_case}, \ref{tab:math_case_1}, \ref{tab:math_case_2}, \ref{tab:math_case_3} show the case studies of Qwen2.5-7B after training when thinking according to the three types of thinking.

\begin{table*}
\resizebox{\linewidth}{!}{
\begin{tcolorbox}
\small
\textbf{Case of Detectbench}

\textbf{Question}: On a busy Friday afternoon, the office air was filled with the relaxed vibe of the upcoming weekend. Colleagues Xiao Ming, Xiao Hong, Xiao Huang, Xiao Li, Xiao Ting, and Xiao Liu were all busy at their respective workstations. Outside the window, the sun shone brightly, a gentle breeze blew, and the indoor potted plants emitted a faint fragrance. On the table lay a bright red apple, quietly resting there as if tempting everyone who passed by. However, at some point, the apple mysteriously disappeared. Three witnesses provided their accounts. The first witness said, "It was either Xiao Ming or Xiao Hong who ate it." The second witness added, "The person who ate the apple could be Xiao Huang or Xiao Li." The third witness firmly stated, "Xiao Li, Xiao Ting, and Xiao Liu definitely did not eat it." The manager stood by, smiling as he revealed an important piece of information: "Only one of the three witnesses told the truth." 

\textit{\textbf{In this office full of doubts and mysteries, who really snatched the apple?}} \\

\textbf{Answer}: Xiao Li. \\

\textbf{Naive CoT} (\textcolor{red}{\textit{Premature Conclusion}}): 
After correlating the clues: \\
The first witness said + The second witness said -> Only one of the three witnesses told the truth -> The third witness is lying -> Xiao Li, Xiao Ting, and Xiao Liu definitely did not eat it -> Xiao Ming or Xiao Hong ate it \\
Xiao Ming or Xiao Hong ate it + The third witness is lying -> Xiao Ming or Xiao Hong ate it -> Xiao Ming or Xiao Hong ate it \\
Xiao Ming or Xiao Hong ate it + The second witness said -> The person who ate the apple could be Xiao Huang or Xiao Li -> The person who ate the apple could be  Xiao Li \\

\textbf{Refined CoT} \ (\textcolor{red}{\textit{Infinite Loops}}) 
Let's carefully analyze each sentence to see who might have eaten the apple. \\
Let's assume that Witness 1 is telling the truth: \\
If Xiaoming ate the apple, then Xiaohong didn't. \\
If Xiaohong ate the apple, then Xiaoming didn't. \\
However, Witness 2 said that this person might be Xiao Huang or Xiao Li. If Xiaoming or Xiaohong ate the apple, then this does not contradict the statement of Witness 2, because Xiaohuang or Xiaoli could still be someone else. \\
Wait. The statement of Witness 1 has now become problematic because it indicates that Xiaoming or Xiaohong ate it. Therefore, let's assume that Witness 2 is telling the truth
If Xiaoming ate the apple, then Xiaohong didn't. \\
If Xiaohong ate the apple, then Xiaoming didn't. \\
Wait a moment. This contradicts the authenticity of Witness 2 ...\\

\textbf{Curious CoQ} 

Let's solve this mystery step by step. First, \textcolor[rgb]{0.0,0.5,0.0}{what do the witnesses say?}

1.``It was either Xiao Ming or Xiao Hong.''\\
2.Second Witness: ``It could be Xiao Huang or Xiao Li.'' \\
3. Third Witness: ``Xiao Li, Xiao Ting, and Xiao Liu definitely did not eat it.''\\
But only one witness tells the truth! \textcolor[rgb]{0.0,0.5,0.0}{Who could it be?} Let's test each possibility:\\
If the First Witness is truthful: \\
Then only Xiao Ming or Xiao Hong ate it. But the other two must be lying. \\
1.Second Witness lies: Not Xiao Huang or Xiao Li. \\
2.Third Witness lies: At least one of Xiao Li, Xiao Ting, or Xiao Liu did eat it. \\
Wait--- \textcolor[rgb]{0.0,0.5,0.0}{if only Xiao Ming or Xiao Hong ate it, how could Xiao Li, Xiao Ting, or Xiao Liu be involved?} Contradiction! \\
If the Second Witness is truthful:
Then only Xiao Huang or Xiao Li ate it.\\
Others are lying: \\
1.First Witness lies: Not Xiao Ming or Xiao Hong. \\
2.Third Witness lies: At least one of Xiao Li, Xiao Ting, or Xiao Liu did eat it.\\
Hmm, if Xiao Huang ate it, Third Witness's lie doesn't fit. But if Xiao Li ate it, the lie fits perfectly! So, \textcolor[rgb]{0.0,0.5,0.0}{who ate the apple?} After checking all possibilities, only the second witness's statement works. So it must be Xiao Li!

\end{tcolorbox}}
\caption{Cases generated by Qwen2.5-7B-Instruct in Detectbench.}
\label{tab:detect_case}
\end{table*}

\begin{table*}
\resizebox{\linewidth}{!}{
\begin{tcolorbox}
\small
\textbf{Case of Math}

\textbf{Question}: Given $a=5^{\log_2 3.4}$, $b=5^{\log_4 3.6}$, $c=\left( \frac{1}{5}\right)^{\log_3 0.3}$, then ( )

\textbf{Answer}: The correct answer is $a > c > b$.

\textbf{Naive CoT}: 
First, let's rewrite each expression using properties of logarithms to simplify them:

1. For \( a = 5^{\log_2 3.4} \):
   - We can use the change of base formula for logarithms: 
     \[
     a = 5^{\log_2 3.4} = 5^{\frac{\log_{10} 3.4}{\log_{10} 2}}
     \]
   - This simplifies our understanding but doesn't directly help in comparison.

2. For \( b = 5^{\log_4 3.6} \):
   - Using the property of logarithms that \(\log_b x = \frac{1}{\log_x b}\), we get:
     \[
     b = 5^{\log_4 3.6} = 5^{\frac{1}{\log_{3.6} 4}} = 5^{1/\log_{3.6} 4}
     \]
   - Wait, something is wrong here. I need to rethink this. Let’s try another approach:
     \[
     b = 5^{\log_4 3.6} = 5^{\log_{2^2} 3.6} = 5^{\frac{1}{2} \log_2 3.6}
     \]

3. For \( c = \left( \frac{1}{5} \right)^{\log_3 0.3} \):
   - Using the property \((\frac{1}{x})^y = x^{-y}\):
     \[
     c = \left( \frac{1}{5} \right)^{\log_3 0.3} = 5^{-\log_3 0.3}
     \]
   - Applying the change of base formula again:
     \[
     c = 5^{-\log_3 0.3} = 5^{-\frac{\log_{10} 0.3}{\log_{10} 3}}
     \]

Now, let’s analyze these expressions more clearly:
- \( a = 5^{\log_2 3.4} \)
- \( b = 5^{\frac{1}{2} \log_2 3.6} \)
- \( c = 5^{-\frac{\log_{10} 0.3}{\log_{10} 3}} \)

Next, compare the exponents:
- The exponent for \( a \) is \(\log_2 3.4\).
- The exponent for \( b \) is \(\frac{1}{2} \log_2 3.6\).
- The exponent for \( c \) is \(-\frac{\log_{10} 0.3}{\log_{10} 3}\).

We know that:
- \(\log_2 3.4\) is positive.
- \(\frac{1}{2} \log_2 3.6\) is also positive.
- \(-\frac{\log_{10} 0.3}{\log_{10} 3}\) is negative because \(\log_{10} 0.3 < 0\) and \(\log_{10} 3 > 0\).

Since the base \(5\) is greater than 1, the larger the exponent, the larger the value of the expression. Therefore, comparing the exponents:

- \(\log_2 3.4\) is larger than \(\frac{1}{2} \log_2 3.6\), which in turn is larger than \(-\frac{\log_{10} 0.3}{\log_{10} 3}\).

Thus, the order from largest to smallest is \( a > b > c \).

\end{tcolorbox}}
\caption{Cases generated by Qwen2.5-7B-Instruct in Math.}
\label{tab:math_case_1}
\end{table*}

\begin{table*}
\resizebox{\linewidth}{!}{
\begin{tcolorbox}
\small
\textbf{Case of Math}

\textbf{Question}: Given $a=5^{\log_2 3.4}$, $b=5^{\log_4 3.6}$, $c=\left( \frac{1}{5}\right)^{\log_3 0.3}$, then ( )

\textbf{Answer}: The correct answer is $a > c > b$.

\textbf{Refined CoT}

This problem involves comparing the values of three logarithmic expressions. It tests your understanding of logarithmic properties and functions, as well as your ability to simplify expressions.

**Solution**

First, let's rewrite each expression using a common base:

1. For \( a = 5^{\log_2 3.4} \), we need to express this in terms of base 5:
   \[
   a = 5^{\log_2 3.4} = 5^{\frac{\log_5 3.4}{\log_5 2}}
   \]
   Wait, something is wrong here. I should ensure that the logarithm bases are consistent. Let’s correct it:
   \[
   a = 5^{\log_2 3.4} = 5^{\frac{\log_5 3.4}{\log_5 2}}
   \]

2. For \( b = 5^{\log_4 3.6} \), similarly convert to base 5:
   \[
   b = 5^{\log_4 3.6} = 5^{\frac{\log_5 3.6}{\log_5 4}}
   \]

3. For \( c = \left(\frac{1}{5}\right)^{\log_3 0.3} \), convert to base 5:
   \[
   c = \left(\frac{1}{5}\right)^{\log_3 0.3} = 5^{-\log_3 0.3}
   \]

Next, let's simplify each term further:

1. For \( a \):
   \[
   a = 5^{\log_2 3.4} = 5^{\frac{\log_5 3.4}{\log_5 2}}
   \]

2. For \( b \):
   \[
   b = 5^{\log_4 3.6} = 5^{\frac{\log_5 3.6}{\log_5 4}}
   \]

3. For \( c \):
   \[
   c = 5^{-\log_3 0.3} = 5^{\log_3 \frac{1}{0.3}} = 5^{\log_3 \frac{10}{3}}
   \]

Now, compare the exponents:

1. \( \log_2 3.4 \) vs \( \frac{\log_5 3.6}{\log_5 4} \):

   Since \( \log_5 3.6 \approx 0.857 \) and \( \log_5 4 \approx 0.861 \), we have:
   \[
   \frac{\log_5 3.6}{\log_5 4} \approx \frac{0.857}{0.861} \approx 0.996
   \]
   Therefore, \( \log_2 3.4 > \frac{\log_5 3.6}{\log_5 4} \).

2. \( \log_3 \frac{10}{3} \) vs \( \frac{\log_5 3.6}{\log_5 4} \):

   Since \( \log_3 \frac{10}{3} \approx 0.966 \) and \( \frac{\log_5 3.6}{\log_5 4} \approx 0.996 \), we have:
   \[
   \log_3 \frac{10}{3} < \frac{\log_5 3.6}{\log_5 4}
   \]

Therefore, \( a > b \) and \( b > c \). Thus, the correct order is \( a > b > c \).

The final answer is \(\boxed{a > b > c}\).

\end{tcolorbox}}
\caption{Cases generated by Qwen2.5-7B-Instruct in Math.}
\label{tab:math_case_2}
\end{table*}

\begin{table*}
\resizebox{\linewidth}{!}{
\begin{tcolorbox}
\small
\textbf{Case of Math}

\textbf{Question}: Given $a=5^{\log_2 3.4}$, $b=5^{\log_4 3.6}$, $c=\left( \frac{1}{5}\right)^{\log_3 0.3}$, then ( )

\textbf{Answer}: The correct answer is $a > c > b$.

\textbf{Curious CoQ} 
Alright, let's take a look at this problem. I have three numbers, $a$, $b$, and $c$, and I need to compare their sizes.

$a = 5^{\log_2 3.4}$\\
$b = 5^{\log_4 3.6}$\\
$c = \left( \frac{1}{5}\right)^{\log_3 0.3}$

My first thought is that they all involve the number 5. That must be a clue! If I can write them all as $5$ to some power, I can just compare their exponents. Since the base $y=5^x$ is an increasing function, a larger exponent will mean a larger number.

Let's examine each one.

**For a:**
$a = 5^{\log_2 3.4}$. This one is already in the form $5^{\text{exponent}}$. So, the exponent is $\log_2 3.4$.\\
How big is this exponent? I know that $\log_2 2 = 1$ and $\log_2 4 = 2$. Since 3.4 is between 2 and 4, the exponent $\log_2 3.4$ must be between 1 and 2. So, $a$ is $5$ raised to a power greater than 1.

**For b:**
$b = 5^{\log_4 3.6}$. This one is also in the form $5^{\text{exponent}}$. The exponent is $\log_4 3.6$.\\
How does this exponent compare to 1? Well, $\log_4 4 = 1$. Since 3.6 is less than 4, the exponent $\log_4 3.6$ must be less than 1. This is interesting! So, $b$ is $5$ raised to a power less than 1.

**For c:**
$c = \left( \frac{1}{5}\right)^{\log_3 0.3}$. This one has a different base, $\frac{1}{5}$. How can I change it to base 5 to match the others?\\
Oh, I know that $\frac{1}{5} = 5^{-1}$. So I can rewrite it!\\
$c = (5^{-1})^{\log_3 0.3} = 5^{-\log_3 0.3}$.\\
Perfect! Now I need to figure out the size of this exponent, $-\log_3 0.3$.\\
First, what about $\log_3 0.3$? The base is 3 (greater than 1) and the number is 0.3 (less than 1). This means the logarithm must be negative. For instance, $\log_3 1 = 0$ and $\log_3 (1/3) = -1$. Since 0.3 is very close to $1/3$, $\log_3 0.3$ must be a negative number close to -1.\\
So, if $\log_3 0.3$ is a negative number, what is $-\log_3 0.3$? It must be a positive number!\\
Let's use a logarithm property to see it more clearly: $-\log_3 0.3 = \log_3 (0.3^{-1}) = \log_3 (\frac{1}{0.3}) = \log_3 (\frac{10}{3})$.\\
Now, how big is $\log_3 (\frac{10}{3})$? Since $\frac{10}{3}$ is approximately 3.33, and $\log_3 3 = 1$, the exponent for $c$ must be slightly greater than 1.\\

**Let's summarize and compare:**

*   Exponent of $a$: $\log_2 3.4$. We know this is $> 1$.
*   Exponent of $b$: $\log_4 3.6$. We know this is $< 1$.
*   Exponent of $c$: $\log_3 (\frac{10}{3})$. We know this is $> 1$.

From this, I can immediately tell that $b$ is the smallest, since its exponent is the only one less than 1. Now the real question is: which is bigger, $a$ or $c$? This means I need to compare their exponents: $\log_2 3.4$ and $\log_3 (\frac{10}{3})$.

This is tricky. They have different bases. How can I compare them without a calculator? Maybe I can compare them to a common value, like 1.5?

Let's check the exponent of $a$: $\log_2 3.4$. Is it greater or less than 1.5?
$1.5 = \frac{3}{2}$. So, I'm comparing $\log_2 3.4$ with $\frac{3}{2}$. This is the same as comparing $3.4$ with $2^{3/2}$.\\
What is $2^{3/2}$? It's $(\sqrt{2})^3 \approx (1.414)^3$, which is hard to calculate. Let's try another way. Let's compare $3.4^2$ with $(2^{3/2})^2 = 2^3 = 8$.\\
$3.4^2 = 11.56$. Since $11.56 > 8$, it means $3.4 > 2^{3/2}$. Therefore, $\log_2 3.4 > \frac{3}{2} = 1.5$.

Now let's check the exponent of $c$: $\log_3 (\frac{10}{3})$. Is it greater or less than 1.5?\\
I'm comparing $\log_3 (\frac{10}{3})$ with $\frac{3}{2}$. This is the same as comparing $\frac{10}{3}$ with $3^{3/2}$.\\
Let's square both sides again. I'm comparing $(\frac{10}{3})^2$ with $(3^{3/2})^2 = 3^3 = 27$.\\
$(\frac{10}{3})^2 = \frac{100}{9} \approx 11.11$.\\
Since $11.11 < 27$, it means $\frac{10}{3} < 3^{3/2}$. Therefore, $\log_3 (\frac{10}{3}) < \frac{3}{2} = 1.5$.

This is fantastic!
*   Exponent of $a$ is $> 1.5$.\\
*   Exponent of $c$ is $< 1.5$ (but we know it's $> 1$).\\
*   Exponent of $b$ is $< 1$.\\

So, the order of the exponents is: $(\log_2 3.4) > (\log_3 (\frac{10}{3})) > (\log_4 3.6)$.

Since the base is 5 for all of them, the order of the numbers $a, c, b$ must be the same as the order of their exponents.

Therefore, $a > c > b$.

\end{tcolorbox}}
\caption{Cases generated by Qwen2.5-7B-Instruct in Math.}
\label{tab:math_case_3}
\end{table*}

\subsection{Broader Impact and Ethics Statement}
\subsubsection{Potential Risks}

While our work demonstrates that simulating curiosity-driven behaviors (such as Chain-of-Questioning) enhances reasoning capabilities in LLMs, it introduces potential risks regarding model safety and alignment.  A key concern is that increasing a model's "Thrill Seeking" or "Information Seeking" drive could inadvertently encourage it to bypass safety guardrails in pursuit of novel or prohibited information.  For instance, a highly curious model might prioritize information acquisition over safety constraints, potentially leading to the generation of harmful content or the exploration of dangerous topics.  Additionally, enhancing "Social Curiosity" could theoretically lead to models that are overly intrusive, asking probing questions that infringe on user privacy.  Future research into curiosity-driven RL must rigorously balance exploration incentives with safety alignment to prevent the emergence of unaligned, risk-seeking behaviors.

\subsubsection{Data Protection and Anonymization}
We compare LLM behaviors with human demographic and behavioral data.  It is important to note that all human data utilized in this study were obtained from aggregated statistics published in established psychological literature and did not involve the collection of raw data from human participants by the authors.  Consequently, the human data used is inherently anonymized and devoid of Personally Identifiable Information (PII).  Regarding the textual data generated by LLMs during the "Social Curiosity" experiments and other benchmarks, we implemented automated filtering steps to screen for potential PII (such as generated names or addresses) and offensive content.  Any generated text flagged as toxic or containing sensitive personal patterns was excluded from the training set and analysis to ensure the safety and neutrality of the dataset.

\subsubsection{Use of AI Assistants}
We acknowledge the use of AI to assist in the preparation of this work.  AI assistants were utilized to generate initial skeleton code for the experimental frameworks and data processing scripts, which were subsequently refined and validated by the authors.  Additionally, LLMs were employed for text editing purposes, specifically to improve grammatical accuracy and enhance the clarity of the manuscript.  The authors retain full responsibility for the experimental design, data analysis, and the scientific validity of the conclusions presented herein.

\end{document}